# A Neural Topic Method Using a Large-Language-Model-in-the-Loop for Business Research

**Stephan Ludwig***

Professor of Marketing

Monash University, Australia

Email: stephan.ludwig@monash.edu

**Peter J. Danaher**

Professor of Marketing and Econometrics

Monash University, Australia

E-mail: peter.danaher@monash.edu

**Xiaohao Yang**

PhD Candidate in Data Science and AI

Monash University, Australia

Email: Xiaohao.Yang@monash.edu

Acknowledgments: The authors thank the Australian Research Council (ARC) for partial financial support during this project (ARC Discovery Project: DP230101490).

*Corresponding author.

3 March 2026

# LX Topic
## An overview

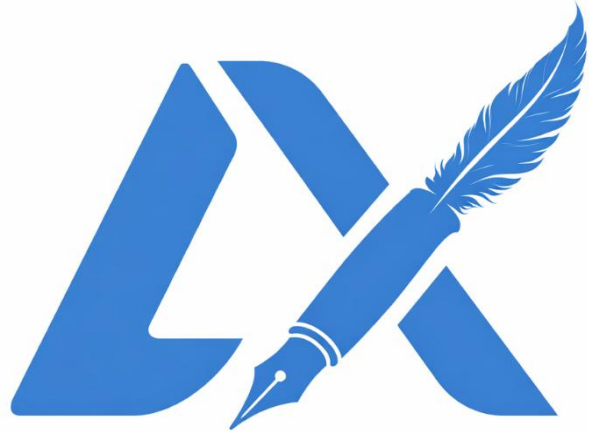

Stephan Ludwig (Professor of Marketing, Monash University, Australia), Peter Danaher (Professor of Marketing and Econometrics, Monash University, Australia), and Xiaohao Yang (PhD candidate Faculty of IT, Monash University, Australia).

Email correspondence relating to the [LXapp](#) should be sent to linguistic.extractor@gmail.com. The LX Topic is a web application that is free to use for academic research purposes. It offers a one-click, research-grade solution for both academic and applied marketing research, enabling scholars and industry professionals to extract empirical, theory-driven insights from unstructured consumer data. We are grateful for the financial support offered by our sponsors without whom this web application would not have been possible. This research was also partially funded by the Australian Research Council Centre of Excellence (project number DP230101490) and the Faculty of Business and Economics at Monash University.

**The official reference to this paper is:**
Ludwig, Stephan, Danaher, Peter J., and Yang, Xiaohao (2026), "**A Neural Topic Method Using a Large-Language-Model-in-the-Loop for Business Research**," Available at arXiv

# A Neural Topic Method Using a Large-Language-Model-in-the-Loop for Business Research

The growing use of unstructured text in business research makes topic modeling a central tool for constructing explanatory variables from reviews, social media, and open-ended survey responses, yet existing approaches function poorly as measurement instruments. Prior work shows that textual content predicts outcomes such as sales, satisfaction, and firm performance, but probabilistic models often generate conceptually diffuse topics, neural topic models are difficult to interpret in theory-driven settings, and large language model approaches lack standardization, stability, and alignment with document-level representations. We introduce LX Topic, a neural topic method that conceptualizes topics as latent linguistic constructs and produces calibrated document-level topic proportions for empirical analysis. LX Topic builds on FASTopic to ensure strong document representativeness and integrates large language model refinement at the topic-word level using alignment and confidence-weighting mechanisms that enhance semantic coherence without distorting document-topic distributions. Evaluations on large-scale Amazon and Yelp review datasets demonstrate that LX Topic achieves the highest overall topic quality relative to leading models while preserving clustering and classification performance. By unifying topic discovery, refinement, and standardized output in a web-based system, LX Topic establishes topic modeling as a reproducible, interpretable, and measurement-oriented instrument for marketing research and practice.

**The Measurement Properties and Research Use of LX Topic**

The increasing availability of unstructured text data, such as customer reviews, social media posts, online forums, and open-ended survey responses, has made topic modelling a central method in business research. Prior work shows that textual content is systematically related to outcomes such as sales, engagement, satisfaction, and firm performance, which has led to the widespread use of text-based measures in empirical models of business research (e.g., Herhausen et al. 2025; Ludwig et al., 2013; Tirunillai and Tellis, 2012,). Topic modelling, in particular, has been used to extract latent themes from large corpora and to operationalize these themes as variables in downstream analyses (Berger et al. 2020; Herhausen et al. 2025; Zhong and Schweidel 2020).

      Despite its popularity, existing topic modelling approaches remain difficult to use as measurement instruments. Classical probabilistic models such as Latent Dirichlet Allocation rely on word co-occurrence patterns and frequently group conceptually unrelated issues into the same topic, limiting interpretability and actionability in applied research contexts (Büschken and Allenby, 2025). Neural topic models improve flexibility and scalability, but their outputs often remain difficult to label and interpret, particularly when topics are intended to serve as explanatory variables in theory-driven empirical work (Ludwig et al. 2025; Zhao et al., 2021). More recent approaches that directly apply large language models to topic modelling can generate readable topic descriptions but are sensitive to prompting, lack methodological standardization (Ziems et al. 2024, Garvey and Blanchard 2025, Oh and Demberg 2025), and often suffer from incomplete topic coverage, weak alignment with document-level representations, and high computational cost when applied to large datasets (Pham et al., 2023; Yang et al., 2024).

LX Topic, where LX stands for Linguistic eXtractor[1], was developed to address these limitations. The model builds directly on recent advances in neural topic modeling and large-language-model-in-the-loop refinement, as formalized in the LLM-ITL framework (Yang et al., 2024), and adapts these advances into a measurement-oriented web application designed for business research. The objective of LX Topic is to extract topics from text data that are interpretable, stable, and suitable for empirical analysis.

The development of LX Topic follows three principles that are consistently emphasized across the methodological and applied materials underlying this work.

- Topics are treated as latent linguistic constructs inferred from the corpus rather than as ad hoc clusters or purely descriptive summaries. Each topic is represented by a list of refined topic words, a short LLM generated topic label and a concise topic description, improving semantic clarity of what each topic means while preserving the underlying topic model structure learned from the data through the application of FASTopic (Wu et al. 2024b, Yang et al., 2024).

- Document-level topic proportions are preserved and weighted for greater clarity in their interpretation. LX Topic produces topic proportion vectors for each document, allowing researchers to quantify how strongly documents attend to different topics. This property is essential for regression analysis, segmentation, longitudinal tracking, and hypothesis testing in business research (Berger et al., 2020).

- LLM-based refinement is applied at the topic-word level rather than at the document level and is integrated using alignment and confidence-weighting mechanisms. This design improves topic coherence while maintaining the overall topic quality of document representations learned by the underlying topic model.

---

[1] The LX app also has the capability to extract emotions and consumer perceptions from unstructured text (see Ludwig et al. 2026).

LX Topic is implemented as a web-based application[2] to ensure accessibility and reproducibility. Users can upload text data (in .csv format), select their preferred number of topics based on some *a priori* notion, or let LX Topic estimate the number of topics resulting in the greatest topic quality, and obtain standardized topic outputs by download link without requiring programming expertise. The system returns topic-level and document-level outputs in formats that can be directly incorporated into empirical models and visual analyses. Any uploaded data and results remain private because LX Topic does not store or share the input data and deletes the output after 7 business days form its drive.

LX Topic is not intended to replace existing topic models. Instead, it integrates neural topic models and large language models into a unified linguistic extraction framework explicitly designed for more interpretable topics and proportions for empirical research. By combining scalable topic discovery with constrained semantic refinement, LX Topic provides researchers with a practical instrument for measuring topic-level attention and content in large text datasets used in business research.

**LX for Topics**

LX Topic conceptualizes topics as latent dimensions inferred from text, where each topic represents a recurring semantic pattern shared across documents in a corpus. Topics are not treated as document clusters or descriptive summaries, but as continuous constructs along which documents vary in the degree to which they allocate attention to a given theme. At the corpus level, LX Topic estimates a shared topic space that defines the semantic structure of the text collection. At the document level, each text is then represented by a vector of topic proportions that quantify the relative (proportionate) presence of each topic within the document. Each topic is associated with a standardized representation consisting of topic words, a topic label, and a topic description, which support consistent interpretation without altering the underlying quantitative measures. Together, the corpus-level topic space and document-level topic proportions form a measurement framework in which topics function as continuous (as a percentage between 0-100%) linguistically derived variables suitable for later use in descriptive overviews, regression analysis, panel models, experiments, and causal inference in business research.

---

[2] https://www.lxapp.net/

**How to work the LX Topic?**

To use LX Topic, users upload a corpus of documents in text format (.csv files only) on the [lxapp.net](lxapp.net) website and specify the column in their .csv file that contains the text that should be analysed. Once uploaded the user can decide to let the LX Topic determine the optimal number of topics (based on topic coherence (TC) and topic diversity(TD)) or self-select a number of topics based on some a priori understanding of their data and context. Once submitted, the system process begins. Text is first cleaned and tokenized, and documents are transformed into representations suitable for topic modelling, following standard pre-processing procedures integrated into the system (Blei et al., 2003; Wu et al., 2024). LX Topic then applies a neural topic model to infer a set of corpus-level topics and corresponding document-level topic proportion vectors, ensuring that each document is represented as a mixture of topics rather than being assigned to a single category. To enhance interpretability, topic-level representations are refined after topic discovery using a large language model in the loop, which operates on topic words rather than full documents. This refinement step improves semantic coherence while preserving the document-topic structure learned from the corpus through alignment and confidence-weighting mechanisms (Yang et al., 2024). The final output of the processing module consists of document-level topic proportions (as a downloadable .csv file), topic-word representations (in a .jsonl file (that can be opened in a notepad), topic labels in the .csv and the .jsonl files), and LLM based topic descriptions (in the .jsonl file), all generated in a standardized format. By separating topic discovery, topic refinement, and document representation within a single processing module, LX Topic ensures that topic-based linguistic measures remain interpretable, comparable across documents, and suitable for empirical analysis in business research. Users are notified by email when their output documents are ready for download via a download link. All uploaded datasets are deleted after processing is complete, the output data is deleted after 7 days and the data is not used for any other purpose[3].

**LX Topic Representativeness**

LX Topic derives its representations directly from the text corpus under analysis, ensuring that topics reflect the content and semantic structure of the data under study (Blei et al., 2003; Wu et al., 2024). Topics are learned at the corpus level using a neural topic model that jointly estimates topic-word distributions and document-level topic proportions, establishing a shared topic space that applies uniformly across all documents.

Each topic is defined as a probabilistic distribution over words, which captures its semantic core and serves as the basis for all subsequent representation. To support interpretation, LX Topic provides four complementary topic-level and document-level outputs.

- Topic words reflect the terms most strongly associated with each topic and define its semantic content (in a .jsonl file format, can be opened in notepad).
- Topic labels provide a concise linguistic identifier that summarizes the dominant theme of the topic (in the .csv file).

---
[3] Please carefully read these terms and conditions (Terms of Use) before you use lxapp.net

- An overview of alternative number of topics and the resulting topic coherence (TC), topic diversity (TD), and overall topic quality (TQ) (in a txt file).
- Topic descriptions offer a brief natural-language characterization of the topic's meaning and context, generated after model training based on the refined topic words and label.

These overviews and representations are designed to support consistent human interpretation while preserving alignment with the underlying topic structure learned from the corpus (Yang et al., 2024).

Topic representations are refined through a post-training process that operates exclusively at the topic level rather than at the document level. For each topic, the most representative topic words are provided as input to a large language model, which generates refined topic words and a concise topic label. This refinement process is constrained through alignment mechanisms that preserve the relationship between the refined topic representation and the original topic-word distribution learned from the corpus. Confidence-weighting limits the influence of the language model when refinement suggestions are uncertain or weakly supported by the data, reducing the risk of semantic drift or over-interpretation (Yang et al., 2024).

As a final step, LX Topic generates topic descriptions for all topics by default. For each topic, the refined topic label and refined topic words are provided as input to llama3.1-8b, which produces a short natural-language description summarizing the semantic content of the topic. This step is applied after model training and calibration and does not affect the inferred topic-word distributions or document-level topic proportions. Topic descriptions are intended to support interpretation, reporting, and communication of results, analogous to category descriptions in dictionary-based linguistic instruments.

By separating topic discovery, topic refinement, and topic description into distinct stages, LX Topic ensures that topics remain empirically grounded in the corpus while being interpretable and reusable. This representation strategy differs from traditional topic modeling approaches that rely on manual inspection and labelling of word lists, as well as from direct large-language-model-based topic generation methods that often lack stable document-level representations. Through standardized yet corpus-adaptive topic representations, LX Topic enables topics to function as consistent and empirically meaningful linguistic measures in business research.

**Development of LX Topic**

The LX Topic model is developed as a topic-based linguistic measurement instrument that builds on recent advances in neural topic modeling and large-language-model-in-the-loop refinement. Formally, given a corpus of documents $D \coloneqq \{d1, \ldots, dN\}$, the model learns a set of corpus-level topics $T \coloneqq \{t1, \ldots, t_K\}$, where each topic $t_k \in \Delta^V$, is a probability distribution over a vocabulary of size $V$, and a corresponding set of document-level topic proportion vectors $Z \coloneqq \{z_1, \ldots, z_N\}$, where each $z_i \in \Delta^K$ represents the mixture of topics present in document $d_i$ (Blei et al., 2003; Wu et al., 2024).

LX Topic adopts a neural topic model as its base component, selected to prioritize document representativeness. Consistent with prior work, the base model learns document-topic representations $z$ and topic-word distributions $t_k$ jointly, ensuring that topic proportions serve as informative and compact representations for downstream tasks such as clustering, classification, and regression (Zhao et al., 2021; Wu et al., 2024). FASTopic (Wu et al., 2024) is used as the base model due to its relatively strongest document representativeness performance on downstream clustering (PN) and classification (ACC) tasks while maintaining computational efficiency (see table 1 for an overview).

**Table 1: Comparison of different Baseline Topic Models**

| Model | Amazon-10Cate PN | Amazon-10Cate ACC | Yelp-5Cate PN | Yelp-5Cate ACC |
|---|---|---|---|---|
| LDA | .242 ± .003 | .336 ± .005 | .435 ± .007 | .655 ± .010 |
| NMF | .127 ± .002 | .272 ± .007 | .284 ± .008 | .636 ± .004 |
| NVDM | .108 ± .002 | .226 ± .009 | .222 ± .006 | .595 ± .006 |
| PLDA | .152 ± .004 | .248 ± .009 | .339 ± .008 | .574 ± .003 |
| SCHOLAR | .230 ± .004 | .334 ± .006 | .492 ± .005 | .662 ± .010 |
| ETM | .221 ± .008 | .330 ± .005 | .442 ± .004 | .687 ± .005 |
| NSTM | .191 ± .005 | .299 ± .005 | .396 ± .011 | .605 ± .012 |
| CLNTM | .214 ± .004 | .319 ± .008 | .474 ± .004 | .644 ± .005 |
| WeTe | .157 ± .002 | .285 ± .003 | .321 ± .006 | .621 ± .017 |
| BERTopic | .210 ± .004 | .406 ± .008 | .424 ± .016 | .712 ± .002 |
| ECRTM | .204 ± .065 | .313 ± .069 | .465 ± .005 | .688 ± .007 |
| FASTopic | **.302 ± .006** | **.440 ± .005** | **.502 ± .003** | **.752 ± .002** |

*Table 1: Document representativeness performance on downstream clustering (PN) and classification (ACC) tasks with K = 50 topics. Reported values are the mean and standard deviation over 5 runs. The best results are highlighted in bold.*

To enhance topic interpretability, LX Topic then integrates large language models into the topic learning process using an LLM-in-the-loop refinement mechanism. After an initial warm-up phase in which the neural topic model learns stable corpus-level topics and document representations, a refinement objective is introduced. For each topic $t_k$, the top-ranked topic words $w'_k$ are provided to a large language model, which generates refined topic words $\mu(w_k, t_k)$ and a concise topic label. The refinement is constrained using an Optimal Transport alignment objective that minimizes the distance between the original topic-word distribution $\mu(w'_k, t_k)$ and the refined distribution, where $u_k$ denotes a uniform distribution over the refined words. The resulting refinement loss is weighted by a confidence score that reflects the reliability of the language model's suggestion, yielding a confidence-weighted refinement objective that is integrated into the overall training loss.

Formally, the training objective combines the neural topic model loss and the refinement loss, with the latter activated only after the warm-up stage. This design ensures that refinement improves semantic coherence without distorting the document-topic distributions learned from the corpus. By operating at the topic level rather than the document level, the refinement process remains computationally efficient and scalable to large corpora. Following topic inference, LX Topic applies a post-estimation calibration to the document-level topic proportions to address a practical measurement issue observed in topic models, namely that topic weights within a document can be highly similar, reducing contrast and interpretability. Let $w_t$ denote the original inferred weight of topic $t$ for a given document and let $w_{min} = min_t\{w_t\}$. LX Topic computes recalibrated topic weights as:

$$w_t^* = \frac{\tanh(w_t - w_{min})}{\sum_{t'} \tanh(w_{t'} - w_{min})}$$

This transformation accentuates relative differences between topics while preserving their order and ensuring that recalibrated weights remain normalized to sum to one. The transformation is applied by default and does not alter topic membership or relative dominance, but improves the practical usability of topic proportions as continuous linguistic measures.

The final, document-level topic proportions, ranging from 0 to 100% for each topic, indicate the degree to which a particular document focuses on each of the topics discussed across all documents (see Table 2 for an illustrative output).

**Table 2: Illustrative LX Topic Output**

| Review Text | Topic Label | | | | | | | | | |
|---|---|---|---|---|---|---|---|---|---|---|
| | **Outdoor Activities** | **Local Produce** | **Customer Service** | **Crime Prevention** | **Retail Experience** | **Culinary Delicacies** | **Luxury Dining** | **City Environment** | **Fashion Style** | **Retail Shopping** |
| Ok, I moved here from LA in July, and was told to go visit the 16th street mall. | .168 | .000 | .121 | .150 | .089 | .028 | .185 | .092 | .072 | .093 |
| OC Transplants Sara and Albert took me here the same day I arrived in Denver. | .199 | .077 | .198 | .217 | .047 | .015 | .000 | .121 | .058 | .067 |
| Hike, bike, walk, run, golf, go to malls like the 16th street mall. | .252 | .071 | .115 | .158 | .089 | .000 | .007 | .200 | .090 | .017 |
| If you want to be asked for money on EVERY fricken street corner, this is the place. | .083 | .105 | .167 | .238 | .057 | .033 | .000 | .221 | .039 | .057 |
| free bus, people watching, variety - YAY! | .260 | .089 | .069 | .121 | .069 | .071 | .007 | .223 | .091 | .000 |

*Table 2: Illustrative output table from LXapp.net showing 13 topics, their respective system generated labels and the respective topic proportions per document.*

**Reliability and Validity of LX Topic**

The reliability, validity, and comparative performance of LX Topic are evaluated using the experimental design, quantitative results, qualitative examples, and visualizations reported in the accompanying method documentation. This section reports the full experimental setup, comparative performance results, representative qualitative outputs, and aggregate visualizations, and references each table and figure explicitly.

*Experimental Setup and Data*

The evaluation is conducted on two benchmark datasets constructed from large-scale consumer text data, as summarized in Table 3. The Amazon-10Cate dataset consists of product reviews drawn from ten product categories, with 15,666 training documents and 4,327 test documents, an average document length of 35 words, and ten ground-truth categories. The Yelp-5Cate dataset consists of reviews from five service-related categories, with 15,998 training documents and 8,474 test documents, an average document length of 112 words, and five ground-truth categories. Text pre-processing includes document cleaning, tokenization, vocabulary construction based on document frequency thresholds, and transformation into bag-of-words representations. These pre-processing steps are integrated into the system, allowing users to upload raw text directly.

**Table 3. Dataset statistics used for experimental evaluation.**

| Dataset | Train Docs | Test Docs | Vocabulary Size | Avg. Doc Length | Categories | Description |
|---|---|---|---|---|---|---|
| Amazon-10Cate | 15666 | 4327 | 3908 | 35 | 10 | Amazon product reviews across ten categories |
| Yelp-5Cate | 15998 | 8474 | 3999 | 112 | 5 | Yelp service reviews across five categories |

*Table 3. reports the datasets and pre-processing statistics used in the experimental evaluation.*

*Quantitative Results and Topic Model Comparison*

Quantitative evaluation focuses on topic coherence, topic diversity, and overall topic quality, where topic quality is defined as the product of topic coherence and topic diversity. Table 4 reports the performance of LX Topic and competing topic models under a fixed number of topics, $K = 50$, across both datasets.

Across both Amazon-10Cate and Yelp-5Cate, LX Topic achieves the highest overall topic quality among all evaluated models, .423 and .450 across Amazon review data and Yelp review data respectively. This improvement is driven by the joint attainment of high topic coherence and high topic diversity, rather than optimizing one dimension at the expense of the other. In contrast, several baseline models exhibit high diversity but lower coherence, or high coherence with substantial redundancy across topics. The results demonstrate that LX Topic produces topics that are internally coherent and well differentiated.

In addition to topic-level evaluation, document-level representational quality is assessed through downstream clustering and classification tasks using topic proportions as input features. The document representativeness results reported earlier in the method section (see Table 1) show that FASTopic-based models outperform alternative topic models on both purity-based clustering metrics and classification accuracy. Because LX Topic builds on FASTopic and integrates refinement in a way that preserves document-topic distributions, these results provide evidence that output interpretability improvements shown in the next section, do not degrade document-level representations.

**Table 4. LX Topic vs. competing topic models**

| Model | Amazon-10Cate | | | Yelp-5Cate | | | Key Reference |
|---|---|---|---|---|---|---|---|
| | TC | TD | TQ | TC | TD | TQ | |
| LDA | .447 ± .007 | .638 ± .021 | .285 ± .013 | .443 ± .006 | .636 ± .024 | .281 ± .008 | Blei, Ng, and Jordan (2003) |
| NMF | .364 ± .002 | .625 ± .002 | .228 ± .001 | .379 ± .008 | .566 ± .006 | .215 ± .006 | Lee and Seung (2000) |
| NVDM | .332 ± .003 | .428 ± .006 | .142 ± .001 | .313 ± .013 | .644 ± .010 | .201 ± .007 | Miao, Grefenstette, and Blunsom (2017) |
| PLDA | .347 ± .007 | .607 ± .007 | .210 ± .005 | .351 ± .013 | .631 ± .024 | .222 ± .012 | Srivastava and Sutton (2017) |
| SCHOLAR | .341 ± .012 | .944 ± .003 | .322 ± .011 | .446 ± .007 | .823 ± .012 | .367 ± .003 | Card, Tan, and Smith (2018) |
| ETM | .417 ± .006 | .728 ± .015 | .303 ± .006 | .387 ± .006 | .564 ± .012 | .218 ± .007 | Dieng, Ruiz, and Blei (2020) |
| NSTM | .462 ± .013 | .613 ± .007 | .283 ± .010 | .384 ± .011 | .234 ± .010 | .090 ± .003 | Zhao et al. (2021b) |
| CLNTM | .335 ± .009 | .942 ± .011 | .316 ± .010 | .459 ± .008 | .769 ± .016 | .353 ± .007 | Nguyen and Tuan (2021) |
| WeTe | .501 ± .008 | .665 ± .030 | .333 ± .010 | .469 ± .008 | .772 ± .048 | .362 ± .027 | Wang et al. (2022) |
| BERTopic | .420 ± .003 | .834 ± .005 | .350 ± .001 | .420 ± .004 | .809 ± .009 | .340 ± .005 | Grootendorst (2022) |
| ECRTM | .306 ± .011 | .913 ± .046 | .279 ± .018 | .313 ± .009 | .931 ± .009 | .292 ± .011 | Wu et al. (2023) |
| FASTopic | .426 ± .008 | .810 ± .010 | .345 ± .005 | .424 ± .008 | .888 ± .011 | .377 ± .011 | Wu et al. (2024b) |
| LLM-ITL | .434 ± .004 | .904 ± .028 | .393 ± .010 | .454 ± .009 | .959 ± .010 | .435 ± .006 | Yang et al. (2025) |
| LX Topic | .465 ± .023 | .909 ± .022 | **.423 ± .024** | .520 ± .014 | .866 ± .034 | **.450 ± .016** | Ludwig, Danaher, and Yang (2026) |

*Table 4: Performance of topic models in terms of topic coherence (TC), topic diversity (TD), and overall topic quality (TQ). The number of topics (i.e., K) is fixed at 50 for all topic models across both datasets. The best performance is highlighted in bold, and the second best is underlined.*

*Interpretive Output and Topic Model Comparison*

Business researchers often need the resulting topics to be interpretable and relevant within their research context. Table 5 presents representative review examples from the Amazon and Yelp datasets along with their top-weighted topics (those topics identified as most important for the respective review text) and the respective topic proportions. Compared to LDA and BERTopic, which are the most commonly used Topic models in business research so far, LX Topic produces top topics for which the underpinning topic words are more semantically coherent and easier to interpret. Noting further, LX Topic automatically assigns a topic label rather than leaving the labelling up to subjective interpretations of what the most frequently occurring 50 words per topic might mean. Moreover, and once again illustrating the better overall topic quality, when documents are examined together with their top-weighted topics, LX Topic yields topics that are more closely aligned with document content, supporting the validity of the extracted topic proportions as indicators of topical emphasis.

While of course only a qualitative example, we encourage researchers to use the topic model and experience the improved corpus level topic interpretability and document-level topical alignment, addressing key limitations of earlier topic modelling approaches.

**Table 5: Illustrative Qualitative Topic Outputs Across Topic Models**

| Document | Model | Topic 1 top words | % | Topic 2 top words | % | Topic 50 top words | % |
|---|---|---|---|---|---|---|---|
| These lashes are really cute !! There's so many pairs and really great quality. I would definitely order again. I like the length and it's the perfect fluff. | LDA | [1] fit size wear small good bit nice love large material | .35 | [2] great good work product quality love easy price recommend excellent | .226 | [50] acid map form stomach store onion shown extract good cell | .000 |
| | BERTopic | [1] shoe comfortable wear heel pair foot love size work fit | .045 | [2] skin taking energy product face pill day vitamin teeth supplement | .039 | [50] lock bike finger safe reliable combination correctly key bag read | .003 |
| | LX Topic | [1] Beauty Application<br><br>Topic Description: The application and use of cosmetic products, such as lipstick, eyeliner, and mascara, to enhance and maintain physical appearance, requiring precise techniques to achieve desired effects without issues like smudging, chipping, or uneven coverage.<br><br>Topic Words: lipstick eyeliner gloss mascara brow coverage foundation liner tattoo smudge | .216 | [2] Fragrance Products<br><br>Topic Description: The fragrance products topic refers to consumer goods that emit pleasant or distinctive odors, encompassing personal care items like perfumes, lotions, deodorants, and diffusers, which are designed to create and maintain a specific scent.<br><br>Topic Words: scent lotion odor perfume diffuser fragrance aroma lavender bath deodorant | .046 | [50] Beverage Preparation<br><br>Topic Description: Beverage preparation involves the processes and methods used to heat and transform various liquids such as coffee, tea, sauce, and soup, as well as cooking methods using equipment like stovetops, microwaves, and ovens to create hot beverage.<br><br>Topic Words: coffee heat hot tea sauce cup cooking microwave soup oven | .004 |
| I just cancelled my membership with these jokes. I was a member for almost a year and they gave me a hard time bringing a guest for 3 days. They have 3 day passes | LDA | [1] gym fitness pool machine equipment member membership tan class staff | .416 | [2] time place thing good day hour make review wanted pretty | .265 | [50] student cleaning daughter clean ryan son red house school labor | .000 |
| | BERTopic | [1] pest control ant company treatment house service home problem mouse | .042 | [2] massage therapist pain spa body relaxing neck tissue deep felt | .039 | [50] mask wearing nose cpap wear cdc kevin cushion policy store | .009 |
| and refused to give me one. The biggest issue about this was I let them know 2 weeks in advance that I was bringing a guest that was coming into town. It wasn't an issue until I actually brought him in and they threw a fit wanting 10 dollars for my cousin to lift there for a day.. The equipment isn't even that good. ALL the barbells are bent and machines are rarely serviced. The staff is cool for the most part. Except Nick!!! He's the reason I left. | LX Topic | [1] Physical Fitness<br><br>Topic Description:<br>A comprehensive physical fitness program encompasses proper nutrition and regular exercising to achieve a significant wellbeing and overall health improvement, typically involving a well-planned workout routine, scientific theory, and professional certification, while allowing for unique and flexible approaches tailored to individual needs.<br><br>Topic Words:<br>nutrition wellbeing fitness proper exercising significant program workout double theory | .101 | [2] Customer Service<br><br>Topic Description:<br>The customer service experience involves interacting with a business to have questions or concerns about a product or order addressed, often resulting in a resolution or resolution attempt, which can impact overall satisfaction with the product and the company.<br><br>Topic Words:<br>review experience customer help need question service trying waited actually | .055 | [50] Fashion Design<br><br>Topic Description:<br>Fashion design encompasses the creation of original and customized garments, accessories, and textiles, involving the manipulation of various materials, such as leather and fabric, to produce visually appealing, functional, and tailored pieces.<br><br>Topic Words:<br>clothing fabric alteration tailor boutique garment fashion designer custo leather | .006 |

*Table 5: Topics and corresponding topic proportions for representative documents from Amazon (top) and Yelp (bottom), as inferred by different topic models. The number of topics is set to 50 for all topic models. Rows reproduce the provided qualitative topic outputs for two documents, showing Topics 1, 2, and 50 and their proportions for each model.*

*Visualization of Topic-Based Linguistic Measures*

For descriptive purposes, LX Topic outputs can be aggregated and visualized. For example, LX Topics returns the key words underpinning each topic with their respective weight for a specific topic. So, as Figure 1 shows, one can chart the respective words and their percentage weighting for each topic. Researchers can further refine their data to a meaningful subset or segment (for example, as illustrated below for 1-star reviews only). This in turn allows them to identify what topics especially matter for a specific segment. So, in this example, LX Topic enables researchers to identify which topics commonly occur in low-rated reviews.

**Figure 1. Illustrative aggregated topic proportions and topic-word visualizations produced by LX Topic for a random subset of 1-star reviews on Yelp.**

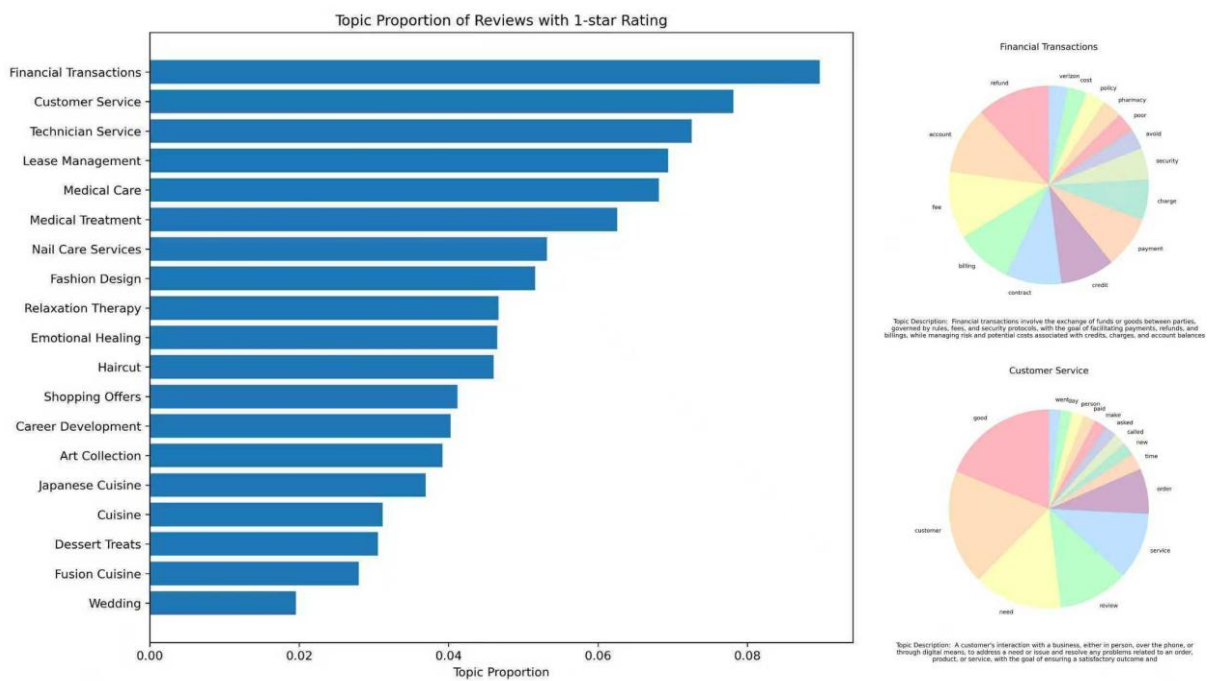

*Figure 1.: Illustration of how LX Topic outputs can be used for visual downstream analysis without additional modeling, supporting exploratory analysis, theory development, and empirical testing.*

**Research and Practical Use of LX Topic**

LX Topic is designed as a practical, easy-to-use, and reliable topic-based linguistic extraction tool that can be applied across datasets, domains, and contexts in business research without requiring users to program or modify the underlying model architecture. Extensibility in LX Topic is achieved through design choices that separate topic modelling and topic refinement, allowing LX topic to adapt to diverse text collections while maintaining standardized outputs. In the current implementation, FASTopic is used as the base model due to its strong document representativeness and low hyperparameter sensitivity, while LLM based topic refinement is performed using a distilled and fine-tuned lightweight language model.

LX Topic also supports both the automatic or user-controlled configuration of the number of topics. Users may specify the desired number of topics for benchmarking or replication purposes, or allow the system to automatically determine an appropriate number of topics based on topic quality, which jointly accounts for topic coherence and topic diversity. This functionality enables LX Topic to be applied to corpora of varying size and complexity without requiring prior domain knowledge or extensive tuning.

From a practical perspective, LX Topic is implemented as a web-based application that provides a one-click interface for topic modelling. Users upload text data in a simple .csv format (max 5 MB, and with all text to be mined in one column), and the system then performs pre-processing, topic modelling, topic refinement, and output generation automatically. The outputs consist of a .csv file that shows all document–topic proportions and topic labels, as well as a separate .jsonl file (that can be opened in a notepad) that shows the topic–word distributions, which are returned in standard file formats and can be directly used for downstream statistical analysis, visualization, or reporting. To further give insights into the decision of the optimal number of topics based on topic quality, LX Topic supplies a .txt file with an overview of different choices of topic number (K) and the corresponding topic coherence scores (TC), topic diversity (TD), and overall topic quality (TQ).

For business researchers, LX Topic outputs are designed to support a wide range of downstream analyses without imposing additional modelling assumptions. Document-level topic proportions can be aggregated, filtered, or compared across subsets or segments of documents defined by metadata such as ratings, categories, or time periods. Topic labels and topic words facilitate interpretation of these analyses, while preserving the quantitative properties of the underlying topic proportions. This combination of extensibility, standardization, and practical usability positions LX Topic as a reusable linguistic extraction instrument rather than a one-off modelling approach.

**Conclusion**

By integrating model development, refinement, calibration, and output generation into a single system, LX Topic provides researchers with a practical measurement instrument for topic-based linguistic analysis. Its design emphasizes standardization, reproducibility, and ease of use, enabling application across domains without requiring extensive technical expertise. Across benchmark datasets, LX Topic demonstrates strong performance in terms of topic coherence, topic diversity, and overall topic quality, while preserving document-level topic proportions that support downstream clustering, classification, and substantive analysis. Qualitative comparisons show that LX Topic produces more coherent and interpretable topics than classical probabilistic models and clustering-based approaches, while maintaining alignment between topics and document content. As such, LX Topic extends the logic of established linguistic extraction tools to topic modelling and positions topic-based measures as reusable and empirically meaningful variables in business and social science research.

Have a look for yourself at lxapp.net, enjoy and good luck with your research!

## Acknowledgements

The authors thank the Australian Research Council (ARC) for partial financial support during this project (ARC Discovery Project: DP230101490). Special thanks to Aaron Guo for continued webapp support.